\def\cvprPaperID{*****}
\def\confName{CVPR}
\def\confYear{2023}

\def\paperTitle{3D Scene Diffusion Guidance using Scene Graphs}

\def\authorBlock{
    Mohammad Naanaa \thanks{Equal contribution} \qquad
    Katharina Schmid \footnotemark[1] \qquad
    Yinyu Nie \\
    
    Technical University of Munich \\
    {\tt\small \{hamudi.naanaa,katharina.s.schmid\}@tum.de}
}

\newif\ifreview 
\newif\ifarxiv 
\newif\ifcamera \newcommand{\cameraready}{\cameratrue}
\newif\ifrebuttal 

\cameraready 

\pdfoutput=1
\documentclass[10pt,twocolumn,letterpaper]{article}
\ifreview \usepackage[review]{cvpr} \fi
\ifarxiv \usepackage[pagenumbers]{cvpr} \fi
\ifrebuttal \usepackage[rebuttal]{cvpr} \fi
\ifcamera \usepackage{cvpr} \fi

\usepackage{graphicx}
\usepackage{amsmath}
\usepackage{amssymb}
\usepackage{booktabs}

\usepackage{times}
\usepackage{microtype}
\usepackage{epsfig}
\usepackage[table,xcdraw]{xcolor}
\usepackage{caption}
\usepackage{float}
\usepackage{placeins}
\usepackage{color, colortbl}
\usepackage{stfloats}
\usepackage{enumitem}
\usepackage{tabularx}
\usepackage{xstring}
\usepackage{multirow}
\usepackage{xspace}
\usepackage{url}
\usepackage{subcaption}
\usepackage{xcolor}
\usepackage[hang,flushmargin]{footmisc}

\ifcamera \usepackage[accsupp]{axessibility} \fi

% ...

\ifarxiv  \fi

\newcommand{\R}[1]{{%
    \textbf{%
        \ifstrequal{#1}{1}{\textcolor{red}{R#1}}{%
        \ifstrequal{#1}{2}{\textcolor{blue}{R#1}}{%
        \ifstrequal{#1}{3}{\textcolor{magenta}{R#1}}{%
        \ifstrequal{#1}{4}{\textcolor{teal}{R#1}}{%
                           \textcolor{cyan}{R#1}%
        }}}}%
    }%
}}

\usepackage{xr-hyper}

\makeatletter
\newcommand*{\addFileDependency}[1]{
  \typeout{(#1)}
  \@addtofilelist{#1}
  \IfFileExists{#1}{}{\typeout{No file #1.}}
}

\makeatother

\usepackage[pagebackref,breaklinks,colorlinks]{hyperref}
\usepackage[capitalize]{cleveref}
\crefname{section}{Sec.}{Secs.}
\crefname{table}{Table}{Tables}
\crefname{figure}{Fig.}{Figs.}

\begin{document}
\title{\paperTitle}
\author{\authorBlock}
\maketitle

\begin{abstract}
Guided synthesis of high-quality 3D scenes is a challenging task. Diffusion models have shown promise in generating diverse data, including 3D scenes. However, current methods rely directly on text embeddings for controlling the generation, limiting the incorporation of complex spatial relationships between objects. We propose a novel approach for 3D scene diffusion guidance using scene graphs. To leverage the relative spatial information the scene graphs provide, we make use of relational graph convolutional blocks within our denoising network. We show that our approach significantly improves the alignment between scene description and generated scene.

\end{abstract}

\section{Introduction}
\label{sec:intro}
Guided 3D scene generation is an emerging research area that has the potential to revolutionize several domains, including computer graphics, virtual reality, and gaming. By allowing users to generate 3D scenes with ease precisely matching an input description, this technology can help streamline the design process, enable more immersive experiences, and facilitate the creation of realistic and complex simulations. Current state-of-the-art methods \cite{paschalidou2021atiss, tang2023diffuscene} already achieve promising results for 3D indoor scene generation but often fall short in accurately aligning the generated scenes with the input description. One possible reason for this limitation lies in the reliance of these models solely on text embeddings of scene descriptions for guidance. To address this issue, we hypothesize that conditioning on scene graphs, which can capture more complex relationships between objects, will yield superior results. With this motivation, we propose a novel diffusion model that generates a 3D scene matrix, encoding a labeled bounding box representation of the scene, guided by scene descriptions encoded as graphs.

The key contributions of the paper include:
\begin{itemize}[itemsep=-2pt]
 \item We present a novel approach for 3D scene diffusion guidance using scene graphs.
 \item Our method includes a novel technique to condition matrix-shaped data on scene graphs leveraging relational graph convolutional networks.
 \item We demonstrate that our approach significantly improves the alignment of the generated scenes with the given conditions.
\end{itemize}

\section{Related Work}
\label{sec:related}

\paragraph{Diffusion Models} Diffusion models, also known as denoising diffusion probabilistic models (DDPMs), have gained significant attention in the field of machine learning.

Diffusion models have been successfully applied to several generative tasks, including natural language processing \cite{li2022diffusionlm}, text-to-speech \cite{kim2022guidedtts}, and text-to-image like \textit{DALL-E} \cite{ramesh2022hierarchical} and \textit{Imagen} \cite{saharia2022photorealistic}, outperforming Generative Adversarial Networks (GANs) on several image synthesis tasks \cite{dhariwal2021diffusion}.
These models provide a powerful framework for probabilistic modeling and have inspired several recent advancements in generative modeling.
In the realm of 3D scenes, diffusion models have received relatively less attention compared to their counterparts in 2D applications. Existing works in the field have primarily focused on generating single objects \cite{luo2021diffusion, zeng2022lion, zhou20213d, hui2022neural, zhang20233dshape2vecset, poole2022dreamfusion}.

\paragraph{3D Scene Synthesis} Synthesizing 3D scenes poses unique challenges due to their higher level of semantics, geometric complexity, and larger spatial extent.
Existing approaches for generating 3D scenes typically rely on various types of generative models, including Variational Autoencoders (VAEs) \cite{purkait2020sgvae}, GANs \cite{yang2021indoor}, and autoregressive models \cite{nie2022learning, wang2021sceneformer, paschalidou2021atiss}. GANs are known for their ability to produce high-quality samples, but they often suffer from limited diversity and poor coverage of different modes in the data distribution. On the other hand, VAEs have better mode coverage but struggle to generate faithful samples that accurately represent the target distribution \cite{xiao2022tackling}. \textit{ATISS} \cite{paschalidou2021atiss} demonstrated the generative ability of autorgressive transformer  architectures, synthesizing new objects conditioned on previously generated ones step by step. While this method allows to capture some inter-object relationships, it still falls short in incorporating complex spatial dependencies between multiple objects. 

For generating 3D indoor scenes with high fidelity and complexity, \textit{DiffuScene} \cite{tang2023diffuscene} formulated the scene synthesis as a diffusion process. The method learns a scene configuration by denoising a set of noisy matrices, where each row encodes object properties. This explicit encoding of diverse scene configurations into the denoising process enables generating faithful object configurations that accurately represent the underlying distribution of scene configurations. The authors also proposed a method to condition the generation process on a textual scene description using extracted word embeddings from a pre-trained \textit{BERT} \cite{devlin2019bert} encoder. While \textit{DiffuScene} achieves state-of-the-art results in unconditional scene synthesis, their guidance method often falls short in accurately aligning the generated scenes with a complex input description, as their approach allows for simple generation guidance from the user's perspective but lacks expressiveness for complex relationships between objects.

\paragraph{Graph Conditioning} A graph is a data structure that captures the relationships (edges) between entities (nodes). Each node in the graph contains associated features, and graphs can be further specialized by introducing directionality to edges or assigning relationship types to edges. The explicit representation of object properties, spatial configurations, and semantic relationships has made graph conditioning effective for generating diverse and contextually consistent data for specific applications.
\textit{HouseDiffusion} \cite{shabani2022housediffusion} achieved competitive results in generating floor plans using graph conditioning with 2D coordinates diffusion. In their approach, an undirected graph is utilized, where nodes represent rooms and edges represent door connections.
Similarly, \textit{GeoDiff} \cite{xu2022geodiff} successfully demonstrated conditional 3D molecular generation by leveraging molecular conformation graphs as guidance. In their setup, a molecular conformation graph is used, comprising nodes representing atoms and edges representing inner atomic bonds.
While these methods have shown promising results with relatively simple graph structures, to the best of our knowledge, no existing work has explored conditioning the diffusion process on graphs with directional edges and specified relation types. This presents an interesting research gap that we aim to address in this paper.

\section{Method}
\label{sec:method}
We propose a novel scene-graph-conditioned DDPM for generating 3D scenes.

\subsection{3D Scene Diffusion}
\label{sec:3d_scene_diff}

\paragraph{Scene Matrix} We represent every scene with a scene matrix - a 2D tensor of fixed size $\mathbf{X} \in \mathbb{R}^{N \times D}$ comprising $N$ objects. Each object $\mathbf{o}_i \in \mathbb{R}^D$ is defined by several attributes including its centroid location $ p \in \mathbb{R}^3 $, three normalized axes $ n \in \mathbb{R}^9 $ and its axis-aligned bounding box size $ s \in \mathbb{R}^3 $. Since the number of objects varies across different scenes, we incorporate an additional 'empty' object and pad it into scenes, thus achieving a fixed number of objects for each scene.
This scene representation allows us to directly retrieve a scene representation with labelled bounding boxes.

\paragraph{Diffusion Process} The forward diffusion process is a pre-defined discrete-time Markov chain in the dataspace $\mathcal{X}$, spanning all possible scenes $\mathbf{X} \in \mathcal{X}$. Given a clean scene matrix $\mathbf{X}_0$ from the underlying distribution $q(\mathbf{X}_0)$, we gradually add Gaussian noise to $\mathbf{X}_0$, obtaining a series of intermediate scene matrix representations $\mathbf{X}_1, ..., \mathbf{X}_T$ with the same dimensionality as $\mathbf{X}_0$, according to a pre-defined increasing noise variance schedule $\beta_1, ..., \beta_T$ (with $\beta_1 < ... < \beta_T$). The joint distribution $q(\mathbf{X}_{1:T}|\mathbf{X}_0)$ of the diffusion process can then be expressed as:

\[ q(\mathbf{X}_{1:T}|\mathbf{X}_0) := \prod_{t=1}^{T}{q(\mathbf{X}_{t}|\mathbf{X}_{t-1})} \]

where the diffusion step at time t is defined as:

\[ q(\mathbf{X}_{t}|\mathbf{X}_{t-1}) := \mathcal{N}(\mathbf{X}_{t}; \sqrt{1 - \beta_t} \mathbf{X}_{t-1}, \beta_t \mathbf{I}) \]

A useful property of diffusion processes is that we can directly sample $\mathbf{X}_{t}$ from $\mathbf{X}_{0}$ via the conditional distribution:

\[ q(\mathbf{X}_{t}|\mathbf{X}_{0}) := \mathcal{N}(\mathbf{X}_{t}; \sqrt{\bar{\alpha_t}} \mathbf{X}_{0}, (1 - \bar{\alpha_t}) \mathbf{I}) \]

where $\mathbf{X}_{t} = \sqrt{\bar{\alpha_t}} \mathbf{X}_{0} + \sqrt{(1 - \bar{\alpha_t})} \epsilon$ with $\alpha_t := 1 - \beta_t$, $\bar{\alpha_t} := \prod_{r=1}^t{\alpha_r}$ and $\epsilon$ is the noise used to corrupt $\mathbf{X}_{t}$.

\paragraph{Generative Process} Following the forward diffusion process, the generative denoising process is parameterized as a Markov chain of learnable reverse Gaussian transitions. Given a noisy scene matrix from a standard multivariate Gaussian distribution $\mathbf{X}_{T} \sim \mathcal{N}(\mathbf{0}, \mathbf{I})$ as the initial state at time step $T$, the generative process goes backwards and corrects $\mathbf{X}_t$ obtaining a slightly less noisy version $\mathbf{X}_{t-1}$ at each time step by using a learned Gaussian transition $p_\theta(\mathbf{X}_{t-1}|\mathbf{X}_{t})$ parametrized by a learnable network $\theta$. By repeating this reverse process until the maximum number of steps $T$, we can reach the final state $\mathbf{X}_{0}$, resembling a denoised scene matrix we aim to obtain. Specifically, the joint distribution of the generative process $p_\theta(\mathbf{X}_{0:T})$ is formulated as:

\[ p_\theta(\mathbf{X}_{0:T}) := p(\mathbf{X}_{T}) \prod_{t=1}^{T}{p_\theta(\mathbf{X}_{t-1}|\mathbf{X}_{t})} \]

with

\[ p_\theta(\mathbf{X}_{t-1}|\mathbf{X}_{t}) := \mathcal{N}(\mathbf{X}_{t-1}; \mu_\theta(\mathbf{X}_{t}, t), \Sigma_\theta(\mathbf{X}_{t}, t)) \]

where $\mu_\theta(\mathbf{X}_{t}, t)$ and $\Sigma_\theta(\mathbf{X}_{t}, t)$ are the predicted mean and covariance, respectively, of the Gaussian $\mathbf{X}_{t-1}$ predicted from $\mathbf{X}_{t}$ by the denoising network $\theta$. For simplicity, we set the variance to a constant $\Sigma_\theta(\mathbf{X}_{t}) := \frac{1 - \bar{\alpha_{t-1}}}{1 - \bar{\alpha_{t}}} \beta_t$, although Song et al. have shown that learnable covariances can improve the generation quality \cite{song2020improved}. We follow Ho et al. \cite{ho2020denoising} and, instead of directly predicting $\mu_\theta(\mathbf{X}_{t}, t)$, we estimate the noise $\epsilon_\theta(\mathbf{X}_{t}, t)$ applied to perturb $\mathbf{X}_{t}$, synthesizing more high-frequent details. In this approach, $\mu_\theta(\mathbf{X}_{t}, t)$ can then be reparametrized by substracting the predicted noise according to Bayes' theorem:

\[ \mu_\theta(\mathbf{X}_{t}, t) := \frac{1}{\sqrt{\alpha_t}} (\mathbf{X}_{t} - \frac{\beta_t}{\sqrt{1 - \bar{\alpha_t}}}\epsilon_\theta(\mathbf{X}_{t}, t)) \]

\paragraph{Training Objective} In our DDPM scheduler implementation, the model can be trained to either predict noise $\epsilon$ or $\mathbf{X}_{0}$ directly, using the reparametrization trick. We refer to Ho et al. \cite{ho2020denoising} for the details of the derivation process.

When predicting $\mathbf{X}_{0}$, we utilize a custom loss function we designed for 3D scene generation. This loss function combines multiple components: (1) L2 loss, (2) a volume loss for 3D bounding boxes, and (3) an aspect ratio loss. The incorporation of these loss components enhances the quality and coherence of the generated scenes: the L2 loss ensures a similarity of bounding box coordinates between generated and ground truth scenes, the volume loss targets natural object volumes given their labels, and the aspect ratio loss preserves natural object proportions.

\[ \mathcal{L}_{\text{total}} = \lambda_1 \cdot \mathcal{L}_{l2} + \lambda_2 \cdot \mathcal{L}_{\text{vol}} + \lambda_3 \cdot \mathcal{L}_{\text{ratio}} \]

For the additive noise prediction setting, the introduced custom losses are not meaningful when computed on noise - neither 3D volume of predicted noise nor its aspect ratio have any distinct meaning - and thus the model is trainable using either L1 or L2 loss.

\subsection{Scene Graph Guidance}
\label{sec:scene_graph_guidance}
We employ scene graphs as a conditioning input for our model to describe the scene to be generated. Scene graphs offer the ability to capture complex relationships between objects. We postulate that this approach is more effective in expressing scene descriptions compared to the prevailing method of solely conditioning on text embeddings. 

The nodes of the scene graph condition correspond to the objects we aim to generate within our scene. To extract the textual information associated with these objects, we utilize a pre-trained FastText encoder \cite{bojanowski2017enriching} to generate text embeddings for their respective labels, which offers the advantage of encoding each label using a single token $ c \in \mathbb{R}^{300} $. 
To establish relationships between the objects, directed edges are employed to connect specific nodes. Thereby, the relationship type associated with each edge specifies the relative spatial positioning of the two respective objects.

\begin{figure}[tp]
    \centering
    \includegraphics[width=0.9\linewidth]{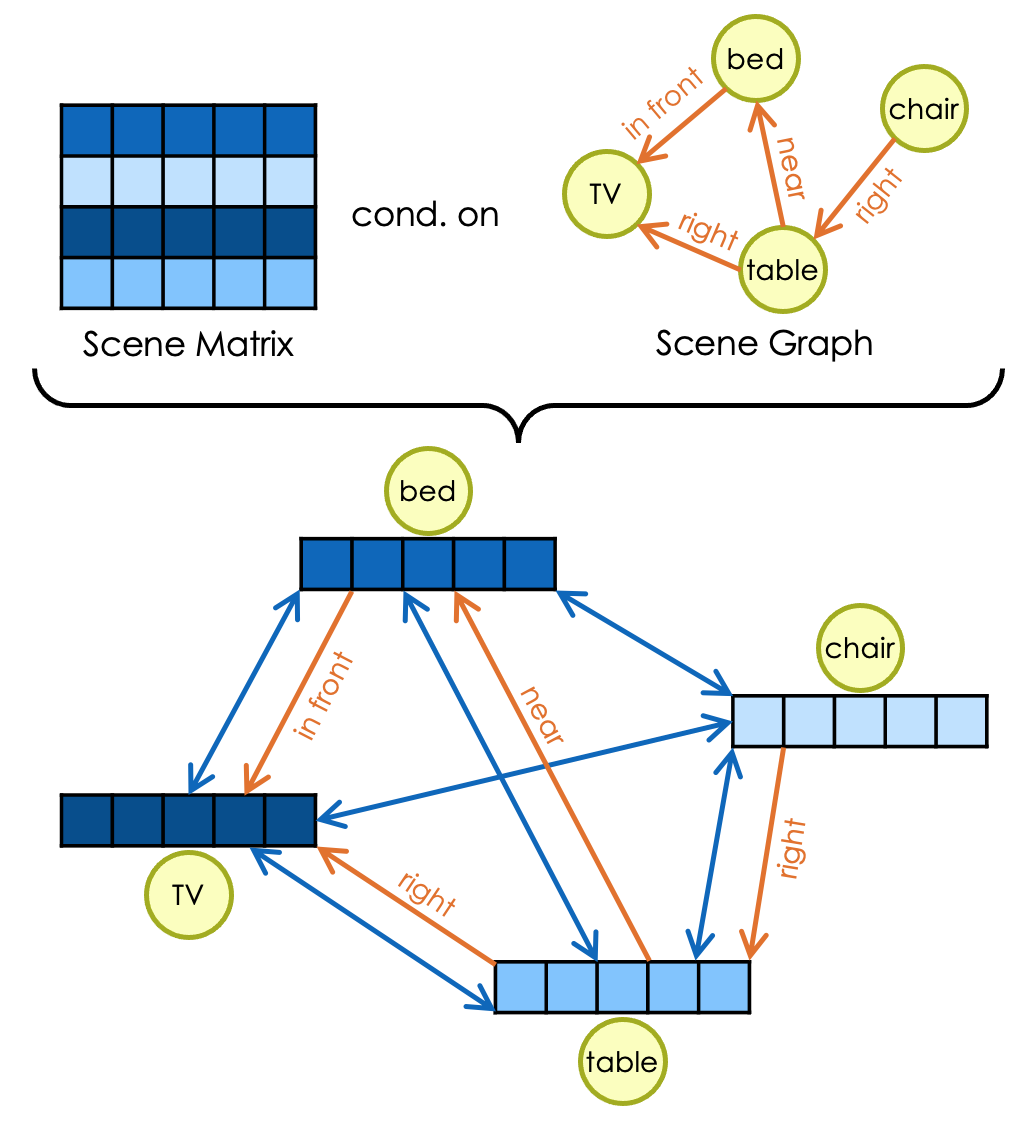}
    \caption{Conditioning of the scene matrix on a scene graph. The scene matrix is converted into a fully-connected graph representation with edges of neutral relationship type (blue).}
    \label{fig:guidance}
\end{figure}

To condition the matrix-shaped scene matrix on a scene graph we follow a series of steps outlined in Fig. \ref{fig:guidance}. First, we convert the scene matrix into a fully-connected graph representation, where each node corresponds to an object embedding. During this conversion process, all edge connections are assigned a neutral relationship type. To leverage the relational information between the objects provided by the scene graph condition, we integrate the edge connections along with their specified relationship type into the existing edges of the graph scene representation.

\begin{figure*}
\begin{center}
\includegraphics[width=0.8\textwidth]{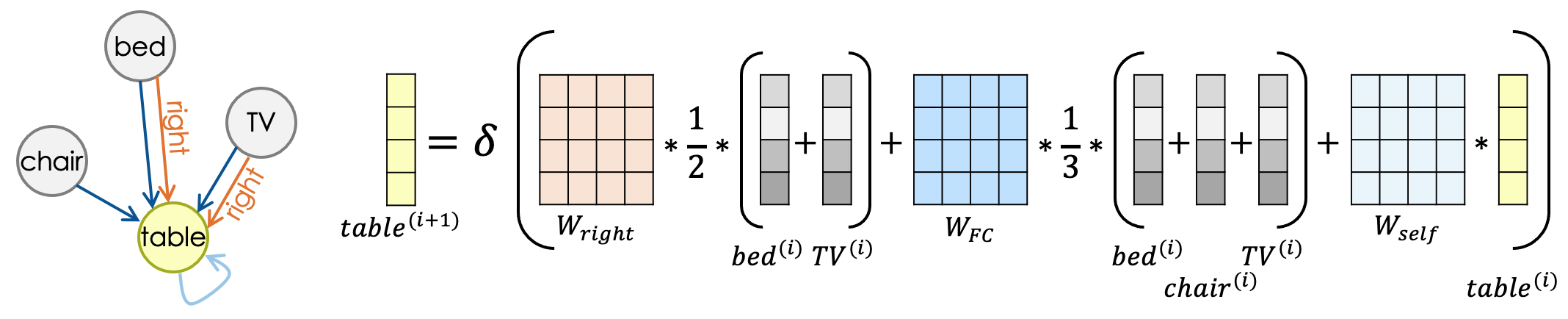}
\end{center}
   \caption{Scene graph processing using Relational Graph Convolutional Networks. The node 'table' is updated based on all nodes connected to it via incoming edges ('chair', 'bed', 'TV'). For each relationship type ('right', 'fully connected') a specific weight matrix is learned. Additionally, a self-connection is employed.}
\label{fig:rgcn}
\end{figure*}

To process the combined graph scene representation, we employ Relational Graph Convolutional Network (RGCN) \cite{schlichtkrull2017modeling} blocks. RGCNs update each node based on the nodes connected to it through incoming edges. This enables the network to learn specific weight matrices for different relationship types describing the relative positioning of the objects in the scene, as shown in Figure \ref{fig:rgcn}. By incorporating spatial and semantic information from edge connections, the network achieves more accurate updates of the objects' bounding boxes. A self-connection further enhances each node's contextual awareness by considering its current state. We found that fully connecting the graph scene representation using a neutral relationship type yields significant benefits. This approach enables the network to effectively update nodes, even when the scene graph condition only provides sparse edge connections between nodes, leading to more robust and comprehensive scene understanding.

\subsection{Architecture}
\label{sec:architecture}

\begin{figure*}
\begin{center}
\includegraphics[width=13cm]{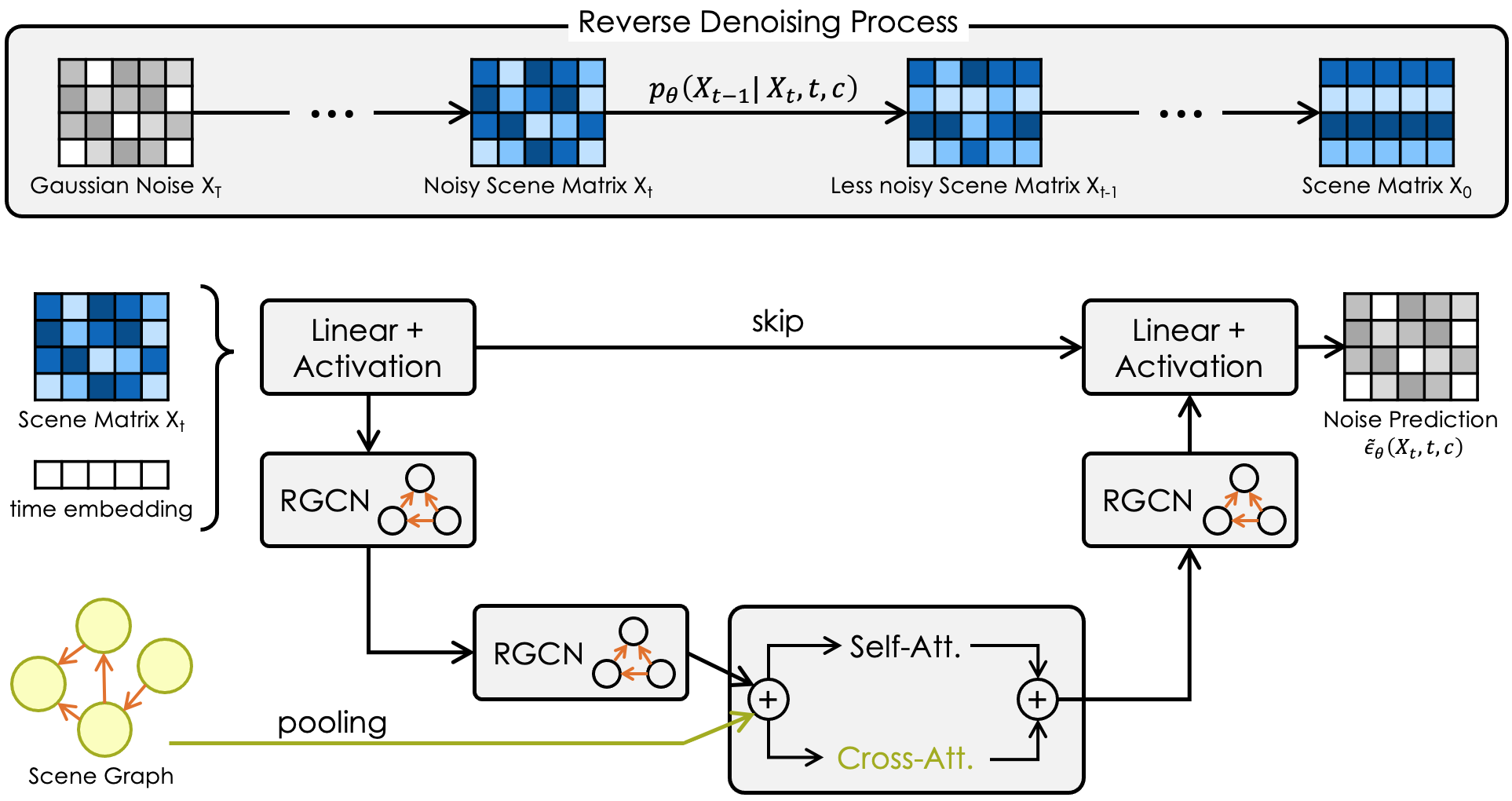}
\end{center}
   \caption{Architecture overview of our denoising network.}
\label{fig:architecture}
\end{figure*}

Fig. \ref{fig:architecture} provides an overview of the architecture we employ. In the initial stage of our denoising network, we incorporate the time step embedding by concatenating it with the input. Linear layers, complemented by a tanh activation function, are then employed to reshape the data according to our specific requirements.
The core components of our denoising network are the RGCN blocks. These blocks enable us to effectively process graph-shaped data, leveraging the inherent relational information. Additionally, we incorporate an attention block in our architecture. Here, we first inject the label information from the condition by adding pooled embedding vectors to the output of the last RGCN block. Subsequently, we simultaneously perform both (1) cross-attention between the condition and the scene matrix to capture label context and (2) self-attention to capture the global context of the scene.
To expedite the training process, we utilize skip connections as a mechanism to facilitate information flow between different layers.

\subsection{Classifier-Free Guidance}
\label{sec:cfg}

In order to enhance the alignment between the desired scene description and our model's output, we adopt the concept of classifier-free guidance, as proposed by \cite{ho2022classifierfree}. Thereby, the model is jointly trained in a conditional and unconditional way by randomly ignoring the condition with a certain probability. Adapting this concept to our problem requires extending the approach to handle relational graph data. To ignore the label data, we follow prior approaches and set the label information to zero \cite{ho2022classifierfree}. Since our scene graph condition also provides relational information, we must additionally mask the respective edge connections, including their specific relationship types. It is worth noting that this approach becomes feasible due to the inherent fully connected neutral edge connections in the unconditioned graph scene representation. This connectedness is essential for meaningful updates in RGCNs, as they rely on edge connections to generate coherent and meaningful outputs in the unconditional case.

The adoption of classifier-free guidance significantly impacted the architecture of our model. To successfully scale the output towards the influence of the condition, it is essential that the model can generate meaningful outputs in an unconditional way. During unconditional training, setting the label information to zero causes the cross-attention layer to produce a zero output. To address this limitation and enable the learning of meaningful outputs in the unconditional scenario, we introduced an alternative information flow. Specifically, we incorporated the cross-attention block in parallel with a self-attention block and added their results, enabling information flow in the unconditional case.

During the inference step we observe that high guidance weights enhance the description-scene alignment. However, this improvement comes at the cost of lower fidelity of data. We attribute this decrease in fidelity to the higher guidance weights causing the predicted values $\mathbf{X}_{0}$ to surpass the bounds of the training data, which is typically in the range of [-1, 1]. To address this issue, we employ dynamic thresholding as a solution, following \cite{saharia2022photorealistic}.

\section{Experiments}
\label{sec:experiments}

Implementing our proposed method, we prepare the required dataset and perform a series of architectural experiments.

\subsection{Dataset}
\label{sec:dataset}
\paragraph{Data Sources} Our method requires a dataset containing both 3D scenes and their corresponding scene graphs for training. We build our experiments on two datasets: The underlying dataset is (1) \texttt{3RScan} \cite{wald2019rio} - a dataset featuring 1482 RGB-D scans of 478 scenes across multiple time steps. Building on top of it, (2) \texttt{3DSSG} \cite{wald2020learning} is a semi-automatically generated dataset that contains semantically rich scene graphs of 3D scenes featuring 1482 samples with 48k object nodes and 544k edges. The authors of the original paper proposed a learned method that regresses a scene graph from the point cloud of a scene. Combining both datasets, we get 3D \textit{scene scans} from \texttt{3RScan} including several objects with ground truth annotations of object labels and bounding boxes, as well as corresponding \textit{scene graphs} from \texttt{3DSSG} describing present objects and their relationships.

\paragraph{Data Filtering} As a real-world point cloud scan dataset, \texttt{3RScan} contains inherent noise and variations, which are further amplified during the semi-automatic scene graph generation conducted by \texttt{3DSSG}. In order to ensure the quality and reliability of our experimental results, we performed a careful filtering process on the dataset, applying several filtering criteria to refine the dataset for our purposes.

Firstly, we filtered out scenes with particularly high noise levels from the original set of 478 environments. Real-world datasets often contain artifacts, bounding box collisions, or incomplete data, which can adversely affect the learning process. By excluding scenes with excessive noise, we aimed to ensure a higher level of consistency and reliability in our dataset. This step was achieved with visual inspection of available ground truth 3D samples.

Furthermore, we reduced the unnecessary complexity of scene graphs by only keeping 23 available spatial relationships, removing visual and functional relationships. This approach allowed us to focus on capturing the \textit{spatial} arrangement of objects, which is crucial for our research contribution goal.

Additionally, to tackle the issue of class imbalance, we opted to retain only the most frequent object categories in the dataset. Real-world scenes often exhibit a wide range of object categories, with some categories being rarely represented. By keeping only the most frequent 51 object categories, we aimed to mitigate the imbalance in the dataset, ensuring a more representative distribution of object types for training our model.

Lastly, following the architectural design of our scene matrix from \ref{sec:3d_scene_diff}, we set the maximum number of representable objects to $N = 20$. We obtained this value empirically as it allows for object-rich scenes but also normalizes outlier scenes containing a lot of small objects that might disproportionately affect the learning process or introduce unnecessary complexity. We filtered out these excessive objects keeping their total per-scene count below 20.

It is important to note that after applying these filtering steps, there were cases where scenes contained no objects or had no relations left. Such cases were also excluded from our final dataset as they provided no meaningful information for our diffusion process.

\paragraph{Final Curated Dataset} As a result of the filtering steps\ifarxiv described in detail in Appendix \ref{sec:appendix_dataset_preprocessing}\fi, we ended up with a curated dataset consisting of 407 aggregated scenes. This refined dataset ensures a more reliable and balanced foundation for evaluating the performance of our conditioning approach for diffusion process in 3D scene generation.

\subsection{Architectural Ablation Studies}
\label{sec:architect_exp}
We design a series of ablation studies to investigate the impact of various design choices on the performance of our scene-graph-conditioned DDPM. We focused on understanding how different architectural components contribute to the generation quality and coherence of the generated 3D scenes. 

\paragraph{Cross-Attention}
The cross-attention block plays a significant role in incorporating label information from the scene graph condition into our model. This enables the model to consider the semantic meaning of the label for the size of the respective bounding box during scene generation. We observed that the cross-attention mechanism is especially useful in ensuring that the generated objects have correct volumes. By incorporating information about object labels, the model can better align the generated shapes with the desired semantics, leading to more coherent and meaningful 3D scenes.

\paragraph{Self-Attention}
Incorporating self-attention in our architecture enhances the overall fidelity of the model. Self-attention helps to address issues such as bounding box collisions, ensuring that the generated objects are spatially distributed in a more realistic manner. Additionally, the self-attention mechanism enables the model to capture global context, which is crucial for maintaining consistency and coherence between different objects in the entire scene.

\paragraph{Time Step Injection}
During our experiments, we explored two approaches for injecting the time step information into the model. The first approach incorporates the time step embedding through an element-wise addition with the input. The second approach involves concatenating the time step embedding with the input at the beginning of the network. We found that the second approach, using concatenation at the beginning, was more effective. This suggests that considering the time step information as an independent factor from the start allows the model to better leverage this temporal information when predicting noise in the current data.

\paragraph{Skip Connection}
We incorporated skip connections in our architecture to facilitate enhanced information flow between different layers. The presence of skip connections improved the gradient flow and accelerated the convergence during training, enabling the model to learn accurate updates for scene objects more easily. This architectural choice helped enhance the stability and overall performance of the model during the scene generation process.

\subsection{Implementation}
\label{sec:implementation}
\paragraph{Loss} Experimentally, we observed marginally better performance when training to predict noise and followed this setting for training our best performing model. For the diffusion process, we tried both linear and cosine schedule and observed a better results with the linear schedule. To increase the stability of denoising training, we normalize the scene matrices to values between $[-1, 1]$.

\paragraph{Classifier-Free Guidance} For the classifier-free guidance, we have found the \textit{condition drop probability} $= 15\%$ and \textit{condition scale} $=3$ \cite{ho2022classifierfree} to achieve the best mode coverage vs. sample fidelity trade-off performance. To balance the effect of high guidance weights on the prediction value distribution, we apply dynamic thresholding with an 85th percentile.

\paragraph{RGCN} For RGCN blocks, we observed $\texttt{mean}(\cdot)$ to be the best aggregation function; regularization with basis-decomposition \cite{schlichtkrull2017modeling} using 4 bases further helped dealing with overfitting on rare relations in RGCN layers.

\paragraph{Training} We train our scene diffusion model on a single Nvidia A10G 24GB GPU with a batch size of 128 for 10000 epochs. We follow the proposed 80/10/10 dataset split by the original dataset authors, slightly rebalancing the splits after the scenes filtering mentioned in \ref{sec:dataset}.  We optimize using Adam with learning rate initialized to $lr = 5e-4$ and weight decay of $1e-4$ with an additional $lr$ reduction on plateau by a $0.8$ factor following $60$ patience threshold.

\section{Evaluation}

We postulated that conditioning the diffusion process on scene graphs leads to enhanced positional alignment since scene graphs provide relative spatial information within the the edges and relationships. We hypothesized that utilizing RGCNs would enable our model to incorporate this relational information into the generated data. 

To evaluate this hypothesis, we present both qualitative and quantitative results, showcasing the effectiveness of our method.

\subsection{Quantitative Results}

\paragraph{Relationship Alignment Score (RAS)} In order to quantify and evaluate the alignment between the generated scenes and the scene graphs used as conditions in our generative process, we propose a new metric called the Relationship Alignment Score (\textit{RAS}). The RAS measures the extent to which the generated scenes capture the relationships specified in the scene graphs.

The calculation of the RAS involves inspecting each relationship $r$ between two objects and determining if the corresponding relation is present in the generated scene or not, assigning it a score $\delta(r)$ of 1 or 0, respectively. To compute the RAS for a complete generated scene $s$, we calculate the average value of scores for all relationships $r \in R_s$ from the scene graph used for guidance. This average value between 0 and 1 indicates the level of alignment between the generated scene and the specified relationships in the scene graph, with higher value denoting a better overall alignment. By computing the RAS for every generated scene $s \in S$ and averaging the scores, we obtain an overall measure of how well the generative process aligns with the specified relationships in the scene graphs.

\[ \text{{RAS}} = \frac{1}{|S|} \cdot \sum_{s \in S} (\frac{1}{|R_s|} \cdot \sum_{r \in R_s} \delta(r)) \]

\[ \delta(r) = \begin{cases} 1, & \text{if relationship } r \text{ holds} \\ 0, & \text{otherwise} \end{cases} \]

The use of the RAS metric allows us to quantitatively assess the ability of our approach to generate scenes which adhere to the specified spatial relationships and semantics defined by the scene graphs. In our experiments, we report the average RAS across all generated scenes to provide a comprehensive evaluation of the alignment with the scene graphs.

\paragraph{Relationship Information Injection} Utilizing the alignment score to assess whether our model is able to effectively make use of spacial relational information provided by the scene graphs, we devised a comparative study between two models. The baseline model considered solely the label information of the condition without incorporating any relational information, whereas the second model was conditioned on the entire scene graph, including edge and relationship information.

\begin{figure}[tp]
    \centering
    \includegraphics[width=\linewidth]{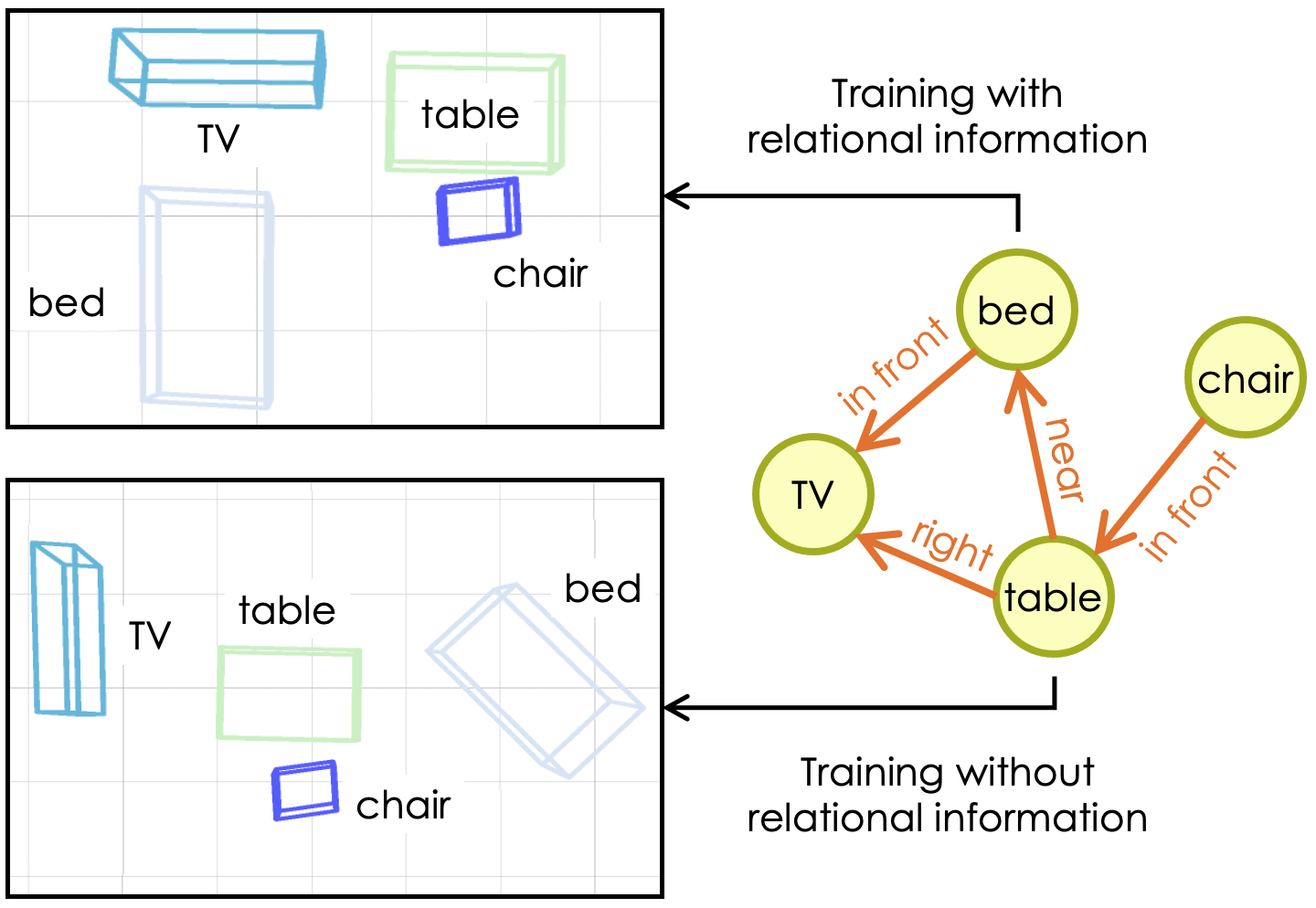}
    \caption{Two synthesized 3D scenes (viewed from top) with the same scene graph condition seen on the right. The top scene was generated by a model incorporating relational information in its RGCN blocks whereas the bottom scene was produced by a model trained without relational information injection. A much better alignment between the scene graph condition and the actual scene is observed by the prior model, matching its higher RAS value.}
    \label{fig:scene_aligned}
\end{figure}

Our findings revealed that the baseline model performed worse and yielded a lower RAS value of \textbf{44\%}, compared to the model trained with relational information injection with an alignment score of \textbf{67\%}, both evaluated on 45 test conditions. This outcome strongly supports our hypothesis that scene graphs offer valuable spatial information which the denoising network can effectively exploit using RGCNs, improving the scene alignment with the condition. This result is also confirmed visually, when comparing the generated scenes of both models for the same condition in Figure \ref{fig:scene_aligned}.

\subsection{Qualitative Results}

\begin{figure*}
\begin{center}
\includegraphics[width=\textwidth]{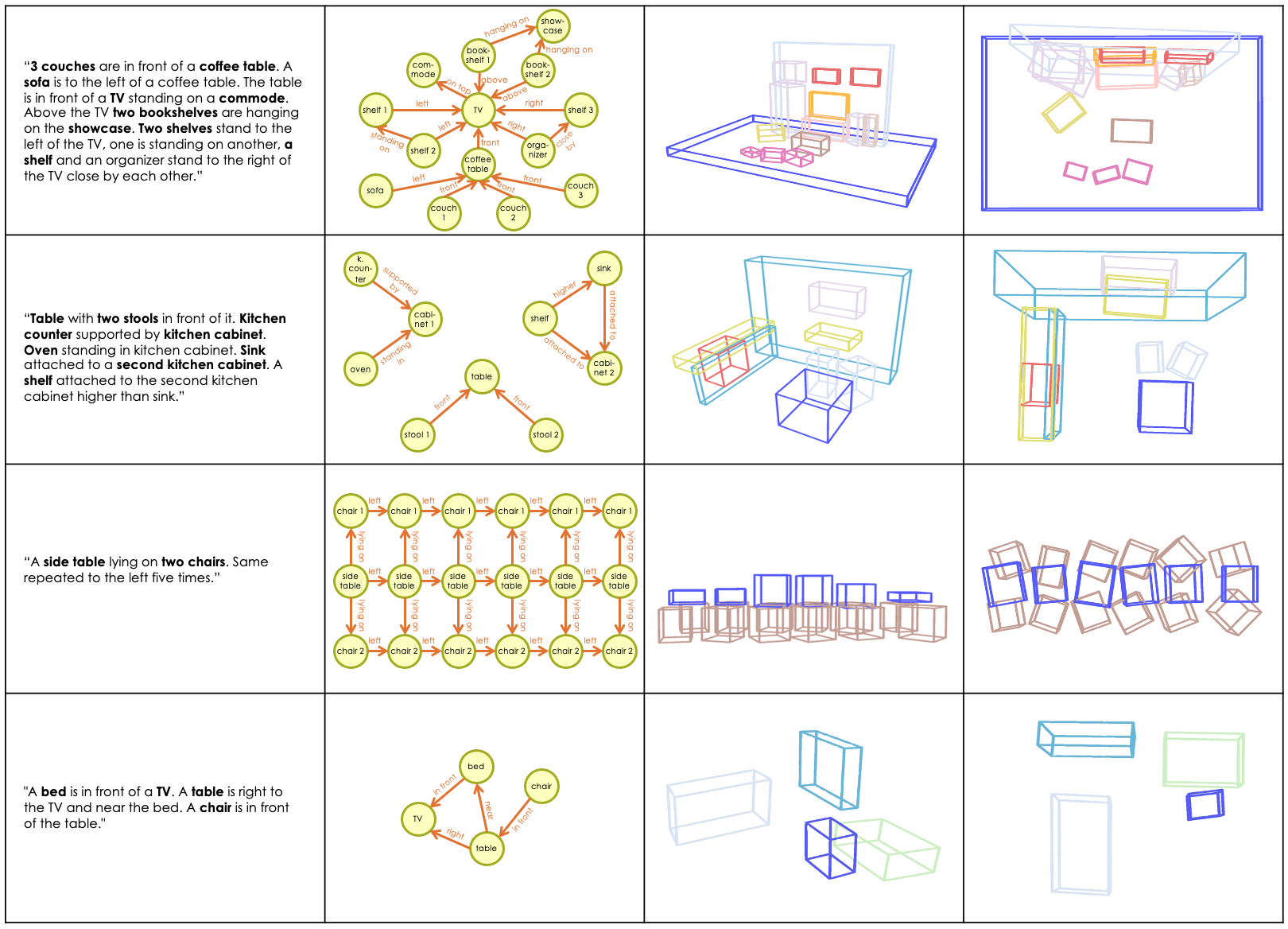}
\end{center}
   \caption{An overview of 3D scene synthesis results. Each row depicts a single result: first column displays a natural language description of a scene, second column shows a corresponding scene graph used as input for the generative process. The remaining two columns depict the synthesized 3D scenes both from the side and top views. The selected results display generative results for (1) very complex, (2) disconnected, (3) repetitive, and (4) simple scene graphs.}
\label{fig:collage}
\end{figure*}

The qualitative evaluation of our proposed method is depicted in Figure \ref{fig:collage}. We observe that the generated scenes align well with the specified scene graphs. For each row in the figure, we provide a natural language description of the scene and the corresponding scene graph that served as the input for the generative process. The subsequent columns display the synthesized 3D scenes, both from the side and top views. We carefully selected results to showcase the effectiveness of our approach across various scenarios, from simple scenes to scenes with higher complexity in object count as well as convoluted relational patterns. Furthermore, even for sparse graph data (e.g. graphs consisting of multiple disconnected subgraphs), our method achieves meaningful results consistent with the description.

The 3D scenes accurately capture the relative spatial positioning of objects and maintain semantic consistency with the given descriptions. Furthermore, the qualitative assessment confirms that the incorporation of relational information using RGCNs significantly enhances the model's ability to align the scenes with the specified relationships and to achieve realistic and contextually appropriate 3D scene synthesis.

Overall, the qualitative evaluation demonstrates that our proposed method is effective in generating 3D scenes that align with the input scene graphs, providing a powerful tool for guided 3D scene synthesis.

\section{Limitations and Outlook}
\label{sec:limitations}

While our method has demonstrated impressive results in creating 3D scenes that conform to the spatial relationship requirements of the condition, it does have certain limitations. 
Firstly, the use of fixed-shaped scene matrices restricts the number of objects within a scene to a pre-defined maximum (e.g., 20). Consequently, our method is unable to synthesize large-scale scenes beyond that limit.
Secondly, in our efforts to validate the effectiveness of our approach in synthesizing scenes which accurately incorporate relative positioning of objects based on the condition, we focused solely on synthesizing labeled bounding boxes. However, for more realistic-looking scenes, we suggest to expand the generated scene matrices to include a shape code $ f \in \mathbb{R}^F $ for each object, as proposed by \cite{tang2023diffuscene}. By leveraging the shape code, along with the bounding box size and object label, we can retrieve the most similar CAD model from a 3D furniture dataset, such as \textit{3D-FUTURE} \cite{fu20203dfuture}, for each object in the generated scene matrix.

The main limitation of our work is caused by the scarcity of paired scene and scene graph data. Our model was trained on a relatively small dataset of less than 400 samples, which significantly limits its ability to generalize. This limitation can only partly be addressed through various data augmentation techniques.
We postulate that the limited number of training samples is responsible for our model's better performance in predicting noise $\epsilon$ as opposed to predicting $\mathbf{X}_{0}$. The model's inability to overfit easily when predicting random noise contributes to this behavior. Regrettably, predicting $\mathbf{X}_{0}$ would have enabled us to employ a more sophisticated loss function, incorporating a volume loss for 3D bounding boxes and an aspect ratio loss as described in \ref{sec:3d_scene_diff}. However, as predicting the noise led to better results on the small dataset, we could only utilize a simple reconstruction loss.
To leverage the advantages of both prediction methods while dealing with the constraints of a limited dataset, we propose an approach to predict both $\mathbf{X}_{0}$ and $\epsilon$, and subsequently merge the predictions by interpolating between predicting $\mathbf{X}_{0}$ directly and predicting via $\epsilon$ \cite{salimans2022progressive}. By doing so, we aim to maintain high variance in the training process while applying a more comprehensive loss function, thus enhancing the realism of the generated scenes.
While previous work by \cite{salimans2022progressive} demonstrated that predicting $\mathbf{X}_{0}$ and $\epsilon$ in parallel is stable and even yields slightly better fidelity scores compared to the standard method of predicting $\epsilon$ only, we leave further experimentation with this method, and its potential for overcoming the challenges posed by a restricted dataset, to future research endeavors.

Finally, in case of an increased availability of training data it is a promising direction to experiment with more complex architectures. Utilizing a larger model holds the potential to achieve significantly improved fidelity scores. As such, an expanded dataset combined with a sophisticated architecture represents an exciting avenue to further elevate the performance of our approach.

\section{Conclusion}
\label{sec:conclusion}
In this paper, we have addressed the challenge of accurately aligning generated scenes with input descriptions in the context of 3D scene generation using denoising diffusion probabilistic models. 
We presented a novel approach using scene graphs with directional edges and specified relationship type to guide the diffusion process, elevating the level of complexity the scene descriptions can encode. To successfully incorporate spatial information from the scene graph condition into the generated scenes we employ relational graph convolutional blocks.
We introduced the \textit{RAS} metric, providing a quantitative tool for evaluating the spatial alignment of generated scenes and consequently enabling the assessment of the effectiveness of our proposed method. 
Throughout our experiments and evaluations, we have demonstrated the efficacy of scene graphs in encoding relative spatial positioning information and the effectiveness of our approach in leveraging this information to achieve significant improvements in scene alignment with the given conditions. 
By effectively addressing the spatial alignment challenge, our novel approach makes a substantial contribution to the field of guided 3D synthesis. We believe that the integration of scene graph information holds immense potential for advancing the capabilities of guided 3D scene synthesis. As such, we hope our work will inspire future research in the realm of scene graph guidance.

{\small
\bibliographystyle{ieee_fullname}
\bibliography{11_references}

\begin{thebibliography}{10}\itemsep=-1pt

\bibitem{bojanowski2017enriching}
Piotr Bojanowski, Edouard Grave, Armand Joulin, and Tomas Mikolov.
\newblock Enriching word vectors with subword information, 2017.

\bibitem{devlin2019bert}
Jacob Devlin, Ming-Wei Chang, Kenton Lee, and Kristina Toutanova.
\newblock Bert: Pre-training of deep bidirectional transformers for language
  understanding, 2019.

\bibitem{dhariwal2021diffusion}
Prafulla Dhariwal and Alex Nichol.
\newblock Diffusion models beat gans on image synthesis, 2021.

\bibitem{fu20203dfuture}
Huan Fu, Rongfei Jia, Lin Gao, Mingming Gong, Binqiang Zhao, Steve Maybank, and
  Dacheng Tao.
\newblock 3d-future: 3d furniture shape with texture, 2020.

\bibitem{ho2020denoising}
Jonathan Ho, Ajay Jain, and Pieter Abbeel.
\newblock Denoising diffusion probabilistic models, 2020.

\bibitem{ho2022classifierfree}
Jonathan Ho and Tim Salimans.
\newblock Classifier-free diffusion guidance, 2022.
\newblock \textit{arXiv:2207.12598}.

\bibitem{hui2022neural}
Ka-Hei Hui, Ruihui Li, Jingyu Hu, and Chi-Wing Fu.
\newblock Neural wavelet-domain diffusion for 3d shape generation, 2022.

\bibitem{kim2022guidedtts}
Heeseung Kim, Sungwon Kim, and Sungroh Yoon.
\newblock Guided-tts: A diffusion model for text-to-speech via classifier
  guidance, 2022.

\bibitem{li2022diffusionlm}
Xiang~Lisa Li, John Thickstun, Ishaan Gulrajani, Percy Liang, and Tatsunori~B.
  Hashimoto.
\newblock Diffusion-lm improves controllable text generation, 2022.

\bibitem{luo2021diffusion}
Shitong Luo and Wei Hu.
\newblock Diffusion probabilistic models for 3d point cloud generation, 2021.

\bibitem{nie2022learning}
Yinyu Nie, Angela Dai, Xiaoguang Han, and Matthias Nießner.
\newblock Learning 3d scene priors with 2d supervision, 2022.

\bibitem{paschalidou2021atiss}
Despoina Paschalidou, Amlan Kar, Maria Shugrina, Karsten Kreis, Andreas Geiger,
  and Sanja Fidler.
\newblock Atiss: Autoregressive transformers for indoor scene synthesis, 2021.

\bibitem{poole2022dreamfusion}
Ben Poole, Ajay Jain, Jonathan~T. Barron, and Ben Mildenhall.
\newblock Dreamfusion: Text-to-3d using 2d diffusion, 2022.

\bibitem{purkait2020sgvae}
Pulak Purkait, Christopher Zach, and Ian Reid.
\newblock Sg-vae: Scene grammar variational autoencoder to generate new indoor
  scenes, 2020.

\bibitem{ramesh2022hierarchical}
Aditya Ramesh, Prafulla Dhariwal, Alex Nichol, Casey Chu, and Mark Chen.
\newblock Hierarchical text-conditional image generation with clip latents,
  2022.

\bibitem{saharia2022photorealistic}
Chitwan Saharia, William Chan, Saurabh Saxena, Lala Li, Jay Whang, Emily
  Denton, Seyed Kamyar~Seyed Ghasemipour, Burcu~Karagol Ayan, S.~Sara Mahdavi,
  Rapha~Gontijo Lopes, Tim Salimans, Jonathan Ho, David~J Fleet, and Mohammad
  Norouzi.
\newblock Photorealistic text-to-image diffusion models with deep language
  understanding, 2022.

\bibitem{salimans2022progressive}
Tim Salimans and Jonathan Ho.
\newblock Progressive distillation for fast sampling of diffusion models, 2022.

\bibitem{schlichtkrull2017modeling}
Michael Schlichtkrull, Thomas~N. Kipf, Peter Bloem, Rianne van~den Berg, Ivan
  Titov, and Max Welling.
\newblock Modeling relational data with graph convolutional networks, 2017.

\bibitem{shabani2022housediffusion}
Mohammad~Amin Shabani, Sepidehsadat Hosseini, and Yasutaka Furukawa.
\newblock Housediffusion: Vector floorplan generation via a diffusion model
  with discrete and continuous denoising, 2022.

\bibitem{song2020improved}
Yang Song and Stefano Ermon.
\newblock Improved techniques for training score-based generative models, 2020.

\bibitem{tang2023diffuscene}
Jiapeng Tang, Yinyu Nie, Lev Markhasin, Angela Dai, Justus Thies, and Matthias
  Nießner.
\newblock Diffuscene: Scene graph denoising diffusion probabilistic model for
  generative indoor scene synthesis, 2023.

\bibitem{wald2019rio}
Johanna Wald, Armen Avetisyan, Nassir Navab, Federico Tombari, and Matthias
  Nießner.
\newblock Rio: 3d object instance re-localization in changing indoor
  environments, 2019.

\bibitem{wald2020learning}
Johanna Wald, Helisa Dhamo, Nassir Navab, and Federico Tombari.
\newblock Learning 3d semantic scene graphs from 3d indoor reconstructions,
  2020.

\bibitem{wang2021sceneformer}
Xinpeng Wang, Chandan Yeshwanth, and Matthias Nießner.
\newblock Sceneformer: Indoor scene generation with transformers, 2021.

\bibitem{xiao2022tackling}
Zhisheng Xiao, Karsten Kreis, and Arash Vahdat.
\newblock Tackling the generative learning trilemma with denoising diffusion
  gans, 2022.

\bibitem{xu2022geodiff}
Minkai Xu, Lantao Yu, Yang Song, Chence Shi, Stefano Ermon, and Jian Tang.
\newblock Geodiff: a geometric diffusion model for molecular conformation
  generation, 2022.

\bibitem{yang2021indoor}
Ming-Jia Yang, Yu-Xiao Guo, Bin Zhou, and Xin Tong.
\newblock Indoor scene generation from a collection of semantic-segmented depth
  images, 2021.

\bibitem{zeng2022lion}
Xiaohui Zeng, Arash Vahdat, Francis Williams, Zan Gojcic, Or Litany, Sanja
  Fidler, and Karsten Kreis.
\newblock Lion: Latent point diffusion models for 3d shape generation, 2022.

\bibitem{zhang20233dshape2vecset}
Biao Zhang, Jiapeng Tang, Matthias Niessner, and Peter Wonka.
\newblock 3dshape2vecset: A 3d shape representation for neural fields and
  generative diffusion models, 2023.

\bibitem{zhou20213d}
Linqi Zhou, Yilun Du, and Jiajun Wu.
\newblock 3d shape generation and completion through point-voxel diffusion,
  2021.

\end{thebibliography}
}

\ifarxiv \clearpage \appendix
\label{sec:appendix}
 \fi

\end{document}